%% file: main.tex
\documentclass[10pt,twocolumn,letterpaper]{article}

\usepackage{cvpr}              
\input{preamble}
\definecolor{cvprblue}{rgb}{0.21,0.49,0.74}
\usepackage[pagebackref,breaklinks,colorlinks,allcolors=cvprblue]{hyperref}

\def\confName{CVPR}
\def\confYear{2026}

\title{Pano3DComposer: Feed-Forward Compositional 3D Scene Generation \\ from Single Panoramic Image}

\author{
    Zidian Qiu \quad \quad Ancong Wu\thanks{Corresponding author: wuanc@mail.sysu.edu.cn} \\
    Sun Yat-sen University \\
    {\tt\small qiuzd@mail2.sysu.edu.cn, wuanc@mail.sysu.edu.cn}
}

\begin{document}

\maketitle

\begin{abstract}
Current compositional image-to-3D scene generation approaches construct 3D scenes by time-consuming iterative layout optimization or inflexible joint object-layout generation. Moreover, most methods rely on limited field-of-view perspective images, hindering the creation of complete $360^\circ$ environments. 
To address these limitations, we design \textbf{Pano3DComposer}, an efficient feed-forward framework for panoramic images. 
To decouple object generation from layout estimation, we propose a plug-and-play Object-World Transformation Predictor. 
This module converts the 3D objects generated by off-the-shelf image-to-3D models from local to world coordinates. 
To achieve this, we adapt the VGGT architecture to \textbf{Alignment-VGGT} by using target object crop, multi-view object renderings and camera parameters to predict the transformation. 
The predictor is trained using pseudo-geometric supervision to address the shape discrepancy between generated and ground-truth objects.
For input images from unseen domains, we further introduce a Coarse-to-Fine (C2F) alignment mechanism for Pano3DComposer that iteratively refines geometric consistency with feedback of scene rendering. 
Our method achieves superior geometric accuracy for image/text-to-3D tasks on synthetic and real-world datasets. It can generate a high-fidelity 3D scene in approximately 20 seconds on an RTX 4090 GPU. 
The project page is available \href{https://qiuzidian.github.io/pano3dcomposer-page/}{here}.
\end{abstract}

\begin{figure}[t]
    \centering
    \includegraphics[width=\columnwidth]{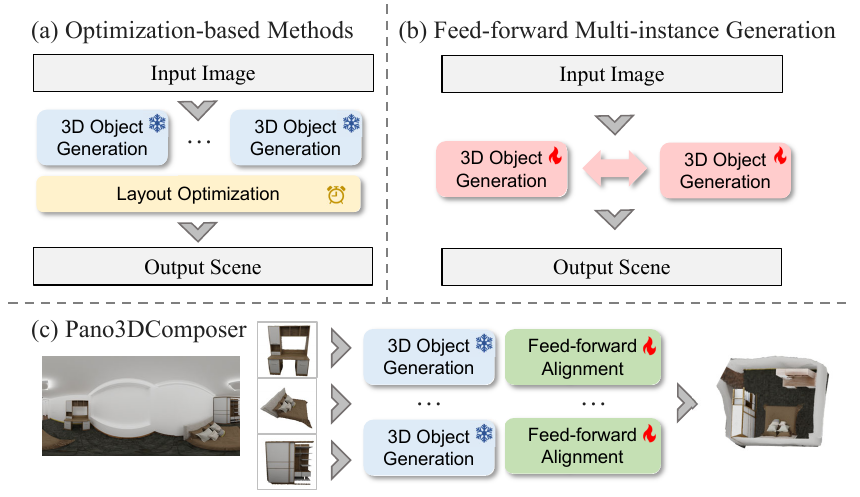}
    \caption{
        Paradigms of compositional 3D scene generation.
    }
    \label{fig:teaser}
\end{figure}

\section{Introduction}
High-quality 3D scene generation underpins applications in VR/AR and digital twins. Despite rapid advances in 3D generation, handling complex multi-object scenes from a single image remains challenging. 
Currently, most advanced image-to-3D pipelines rely on perspective images, which suffer from a limited field-of-view. Conversely, panoramic images overcome this restriction by offering rich spatial context of the entire environment, enabling the generation of geometrically complete $360^\circ$ 3D scenes.

Existing 3D scene generation methods primarily fall into the following categories.
Feed-forward scene understanding methods \cite{nie2020total3dunderstanding,zhang2021im3d,liu2022instpifu,dahnert2024coherent} jointly predict layout, object geometry and poses using encoder–decoder architectures, which are efficient for inference but limited by lack of precise 3D mesh supervision and out-of-distribution generalization.
Feed-forward multi-instance generative models \cite{huang2025midi,meng2025scenegen} extend single-object generators to jointly synthesize multiple instances and the layouts, often requiring costly fine-tuning.
Compositional optimization-based pipelines \cite{Zhou2024GALA3D,zhou2024layoutyour3d,dong2025hiscene,gu2025artiscene,yang2024holodeck,ardelean2025gen3dsr,hu2025flashsculptor,li2024dreamscene} separate asset generation from pose/layout optimization. Unfortunately, these methods usually rely on time-consuming iterative optimization processes, making it difficult to meet efficiency requirements. 
The above discussed methods trained on perspective images are not directly applicable to equirectangular panoramas, because panoramic images exhibit severe distortion, non-uniform sampling, and view-dependent distance/angle foreshortening.
Only a few approaches \cite{zhang2021deeppanocontext,dong2024panocontext} focus on addressing panoramic images, but these methods are limited to generating meshes without textures that cannot compose a render-ready 3D scene.

To overcome the limitation of time-consuming optimization and inflexible joint object-layout generation of existing methods, we design Pano3DComposer, a modular feed-forward framework for compositional 3D scene generation from a single panorama, as shown in Figure~\ref{fig:teaser}.
The core modules of the framework are the 3D Object Generator and the Object-World Transformation Predictor, which decouple object generation from layout estimation.

For object generation, we first segment and project each instance crop from the panorama into the perspective domain, mitigating panoramic distortion. An off-the-shelf 3D Object Generator then produces high-quality meshes or 3D Gaussian splats \cite{kerbl20233dgs} from this localized input.

To position the generated object in the scene, we formulate the object-to-world transformation as a cross-coordinate geometric mapping problem.
The Object-World Transformation Predictor learns the conversion from object renderings of the generated object to the cropped input.
To achieve this, we adapt the VGGT \cite{wang2025vggt} architecture to Alignment-VGGT, which takes multi-view object renderings, target object crop and their corresponding camera parameters as input to estimates object-to-world rotation, translation, and anisotropic scale in a single feed-forward pass.
Furthermore, to handle shape discrepancies between generated and ground-truth (GT) objects, the transformation predictor is trained using pseudo-geometry supervision distilled from differentiable optimizers, instead of GT mesh supervision. 

To handle inputs from unseen domains, we further introduce the Coarse-to-Fine (C2F) alignment mechanism that iteratively refines the geometric consistency of each object using feedback from the current scene rendering.


The feed-forward nature of our framework ensures efficient inference, while its modular design guarantees flexibility that components can be trained independently, and off-the-shelf 3D object generator can be integrated without training.
Our contributions are summarized as follows:
\begin{itemize}[leftmargin=1.2em]
    \item A plug-and-play Object-World Transformation Predictor module based on the Alignment-VGGT architecture enables efficient alignment of a generated 3D object with the target panoramic scene rendering in a forward pass.
    \item A coarse-to-fine alignment mechanism that progressively improves object-to-scene alignment without gradient-based optimization.
    \item Extensive experiments on synthetic and real scenes demonstrate superior geometric accuracy and inference efficiency compared to state-of-the-art methods.
\end{itemize}

\begin{figure*}[t]
    \centering
    \includegraphics[width=0.95\textwidth]{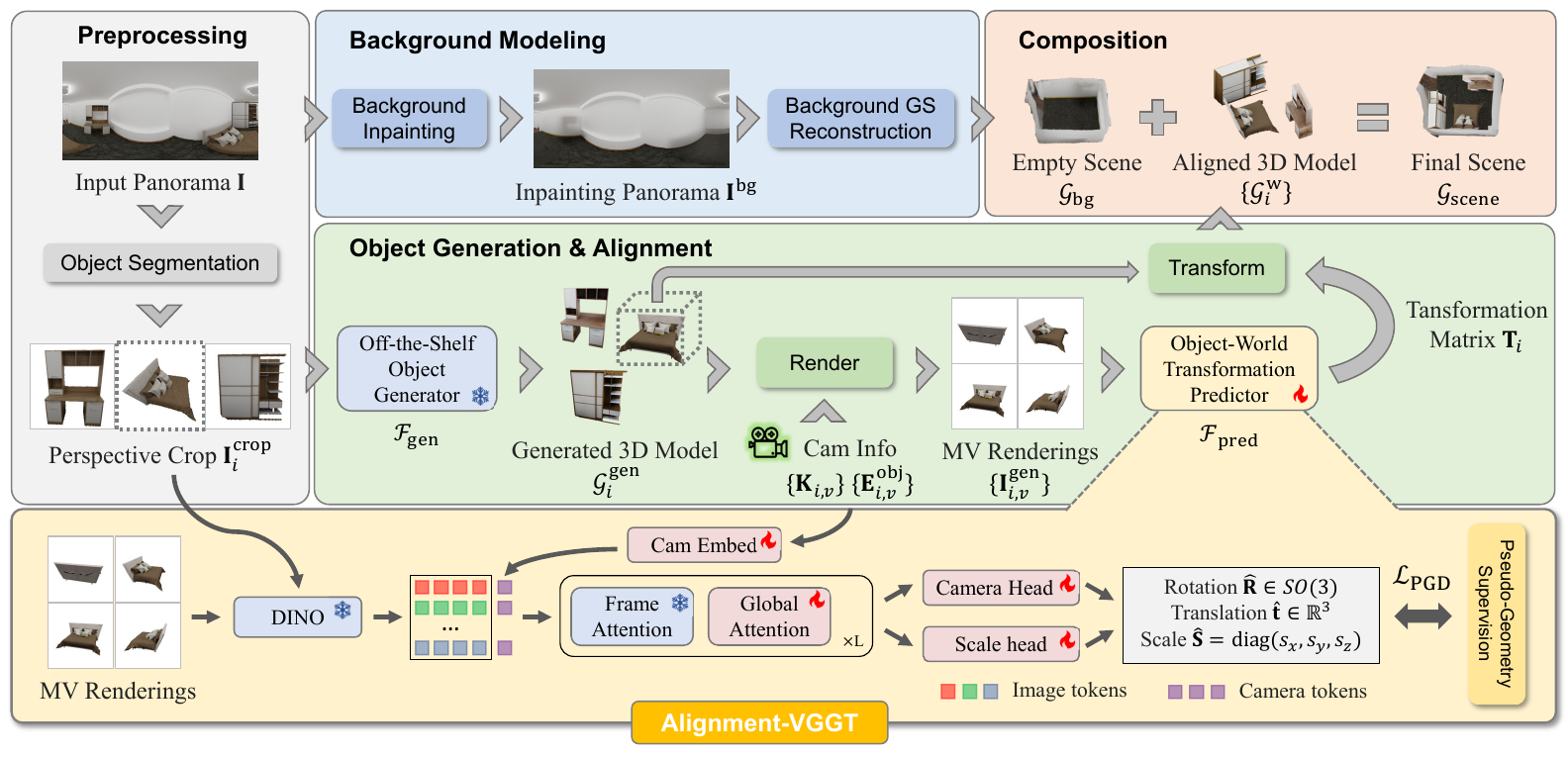}
    \vspace{-0.2cm}
    \caption{
        \textbf{Overview of Pano3DComposer.} 
        The framework takes a panoramic image $\mathbf{I}$ as input and generates a 3D scene $\mathcal{G}_{\mathrm{scene}}$ through four stages: 
        (\textbf{i}) Preprocessing, 
        (\textbf{ii}) Object Generation \& Alignment, 
        (\textbf{iii}) Background Modeling, and 
        (\textbf{iv}) Composition.
    }
    \label{fig:framework}
\end{figure*}

\section{Related Work}
\subsection{Text/Image-to-3D Object Generation}

Diffusion-driven 3D generation has advanced rapidly. For text-to-3D, DreamFusion \cite{poole2022dreamfusion} introduced Score Distillation Sampling (SDS), enabling 3D synthesis from 2D diffusion models \cite{ho2020diffusion}. Follow-ups \cite{lin2023magic3d} adopt more efficient 3D representations such as 3D Gaussian Splatting (3DGS) \cite{kerbl20233dgs} to improve quality and efficiency. Recent work further leverages text-to-point-cloud initialization and human priors \cite{yi2024gaussiandreamer,liang2024luciddreamer}, or structured noise for variational 3DGS \cite{li2023gaussiandiffusion}.
For image-to-3D, methods adapt 2D diffusion models for multi-view synthesis \cite{liu2023zero123,liu2023syncdreamer,shi2023mvdream,shi2023zero123++} and accelerate single-view reconstruction with feed-forward networks such as LRM \cite{hong2023lrm} and LGM \cite{tang2024lgm}. Asset-level training on 3D models further boosts geometric fidelity, as demonstrated by CLAY \cite{zhang2024clay} and TRELLIS \cite{xiang2025trellis}.
However, these approaches primarily focus on single objects and remain limited in multi-object composition. In contrast, we extend single-object generators to compositional scene synthesis via a plug-and-play transformation predictor that aligns multiple objects within $360^\circ$ panoramic views.

\subsection{3D Scene Generation}
\noindent\textbf{Feed-forward 3D scene generation.}
Early methods such as Total3D \cite{nie2020total3dunderstanding} jointly estimate layout, object poses, and shapes from a single image; follow-ups including Im3D \cite{zhang2021im3d} and InstPIFu \cite{liu2022instpifu} adopt implicit representations for indoor reconstruction, or apply 3D diffusion conditioned on input images \cite{dahnert2024coherent}. Extensions to panoramas such as DeepPanoContext \cite{zhang2021deeppanocontext} and PanoContext-Former \cite{dong2024panocontext} recover room layouts and object geometry from a single panoramic image. Another line of research extends large single-object generators to multi-object generation, including MIDI \cite{huang2025midi} and SceneGen \cite{meng2025scenegen}. CAST \cite{yao2025cast} directly predict alignment parameters in a single forward pass. However, these approaches suffer from tight coupling between the object generation module and the alignment module, making it difficult to adopt a plug-and-play design that would allow flexible switching among different object generation models. They also suffer from high training costs and limited handling of panoramic distortions.
Unlike these methods, 
our approach decouples object generation from spatial alignment, enabling flexible integration of any 3D object generator while efficiently handling panoramic distortions through perspective projection.

\noindent\textbf{Compositional 3D scene generation.}
Compositional pipelines decouple object generation and layout optimization, often aided by LLMs for layout planning, exemplified by GALA3D \cite{Zhou2024GALA3D} and LayoutYour3D \cite{zhou2024layoutyour3d}. Several approaches optimize object poses via differentiable rendering or depth alignment, including REPARO \cite{han2025reparo} and Gen3DSR \cite{ardelean2025gen3dsr}, while others retrieve and compose objects from 3D databases, e.g., Holodeck \cite{yang2024holodeck}. Recent efforts, represented by HiScene \cite{dong2025hiscene} and ArtiScene \cite{gu2025artiscene}, advance amodal completion and object generation quality. 
However, they are significantly affected by occlusions in RGB input and inaccuracies in estimated depth, which interfere with pose optimization and prevent precise alignment with the input image. Moreover, most pipelines either rely on slow iterative optimization, and failing to address the unique challenges of panoramic inputs. Our method overcomes these limitations by a feed-forward transformation predictor trained with pseudo-geometry supervision, achieving efficient alignment specifically for panoramas.


\section{Pano3DComposer}
\label{sec:method}
In this section, we introduce Pano3DComposer, a feed-forward modular framework for fast generation of geometrically complete $360^\circ$ environments from panoramic images. 

\subsection{Overall Framework}
\label{sec:overall}
Given a equirectangular panorama \(\mathbf{I} \in \mathbb{R}^{H \times W \times 3}\), we aim to reconstruct a compositional 3D scene consisting of a set of objects \(\{\mathcal{G}_i^\text{w}\}_{i=1}^N\) and background \(\mathcal{G}_{\mathrm{bg}}\) in the world coordinate system. The overall rendering of \(\{\mathcal{G}_i^\text{w}\} \cup \mathcal{G}_{\mathrm{bg}}\) should be consistent with \(\mathbf{I}\) both photometrically and geometrically.

Our framework consists of four stages: (i) Preprocessing, (ii) Object Generation and Alignment, (iii) Background Modeling, and (iv) Composition, as shown in Figure~\ref{fig:framework}.
First, the preprocessing module takes the panoramic image $\mathbf{I}$ as input and outputs a set of distortion-free perspective crops $\{\mathbf{I}_i^{\mathrm{crop}}\}_{i=1}^N$ for each detected object.
Then, each crop is fed into the 3D object generator $\mathcal{F}_{\mathrm{gen}}$ to produce 3D object asset $\mathcal{G}_i^{\mathrm{gen}}$ represented in the object's local coordinate system.
The core module Object-World Transformation Predictor $\mathcal{F}_{\mathrm{pred}}$ learns to predict the coordinate transformation $\mathbf{T}_i=\mathcal{F}_{\mathrm{pred}}(\mathbf{I}^\text{crop}_i,\mathcal{G}_i^{\mathrm{gen}})$. The transformation $\mathbf{T}_i$ converts asset $\mathcal{G}_i^{\mathrm{gen}}$ from local to world coordinate system, in order to obtain $\mathcal{G}_i^{\mathrm{w}} = \{ \mathbf{T}_i\,\mathbf{p} \mid \mathbf{p}\in\mathcal{G}_i^{\mathrm{gen}}\}$ that is aligned with the input panorama.
The background modeling module reconstructs a 3D scene representation $\mathcal{G}_{\mathrm{bg}}$ from the inpainted image $\mathbf{I}^{\mathrm{bg}}$.
Finally, the composition module fuses $\mathcal{G}_i^{\mathrm{w}}$ and $\mathcal{G}^{\mathrm{bg}}$ to obtain a geometrically complete 3D scene.

\subsection{Object Generation and Alignment}
\label{sec:object}

\subsubsection{3D Object Generator}
\label{sec:generator}
\noindent\textbf{Preprocessing.} Given an equirectangular panoramic image \( \mathbf{I} \in \mathbb{R}^{H \times W \times 3} \) that encodes a full \(360^\circ\) view of the scene, each pixel corresponds to a direction defined by longitude and latitude angles \((\theta, \phi)\). 
We first apply open-vocabulary 2D foundation models on the panoramic image, e.g., SAM \cite{kirillov2023sam}, 
to extract masks \(\{\mathbf{M}_i\}_{i=1}^N\) for each object.
Let $\Pi_{\text{persp}}(\cdot;\theta,\phi,\alpha)$ denote a perspective projection operator parameterized by longitude $\theta \in [-\pi,\pi)$, latitude $\phi \in [-\tfrac{\pi}{2},\tfrac{\pi}{2}]$, and field-of-view $\alpha \in (0,\pi)$.
For each object, We use its corresponding  $(\theta_i,\phi_i,\alpha_i)$ to transform the masked object from panorama coordinates to the distortion-free perspective crops \(\{\mathbf{I}_i^{\mathrm{crop}}\}_{i=1}^N\) by
\begin{equation}
\mathbf{I}^\text{crop}_i = \Pi_{\text{persp}}\!\big(\mathbf{I} \odot \mathbf{M}_i;\, \theta_i,\phi_i,\alpha_i\big).
\end{equation}

\noindent\textbf{3D object generation.} With the distortion-free perspective crops $\mathbf{I}^\text{crop}_i$, any off-the-shelf image-to-3D method can be employed to reconstruct the object as a mesh or a set of Gaussian splats. We denote the generated object by ${\mathcal{G}}_i^\mathrm{gen}$, of which the 3D point set of $\mathcal{G}_i^{\mathrm{gen}}$ is denoted by $\mathcal{P}_i^\mathrm{gen}=\{(x_j^\mathrm{obj}, y_j^\mathrm{obj}, z_j^\mathrm{obj})\}_{j=1}^M$ in the local coordinate system.

We use TRELLIS \cite{xiang2025trellis} as the default generator due to its high-fidelity geometry on mildly occluded instances; for severe occlusion we optionally adopt amodal completion (e.g., Amodal3R \cite{wu2025amodal3r}) to improve geometric completeness.

\subsubsection{Object-World Transformation Predictor}
\label{sec:predictor}
\noindent\textbf{Cross-coordinate geometric mapping problem.} The core challenge is accurately placing each generated object $\mathcal{G}_i^{\mathrm{gen}}$ by estimating the transformation $\mathbf{T}_i$ (rotation, translation, scale) to convert the object from its local coordinates to world coordinates for alignment with panorama $\mathbf{I}$.

A straightforward approach is to perform 3D alignment between the generated object and the scene geometry estimated from the monocular panorama. However, accurately reconstructing 3D geometry from a single panoramic image is still very difficult, leading to alignment errors.
To avoid this limitation, we shift the task from the challenging 3D space to more robust 2D image space. We represent the 3D asset using multi-view renderings that captures geometry and texture details, and then seek correspondence with the target perspective object crop from the panorama. 

\noindent\textbf{Preliminaries of VGGT.} Recently, visual-geometry foundation models \cite{wang2025vggt,keetha2025mapanything} have achieved remarkable success. 
These models, such as Visual Geometry Grounded Transformer (VGGT) \cite{wang2025vggt}, take multi-view RGB images as input and predict a set of 3D attributes in a forward pass, including camera parameters, depth maps, and point clouds.

Current foundation models are typically restricted to processing multi-view inputs captured from the same 3D scene that share the same coordinate system and camera parameters. However, the cross-coordinate geometric mapping problem involves input images captured by different camera parameters and represented in different coordinate systems.

\noindent\textbf{Alignment-VGGT.} 
To construct the Object-World Transformation Predictor $\mathcal{F}_{\mathrm{pred}}$, we introduce Alignment-VGGT, an adaptation of the VGGT architecture with specialized input structure and output heads to bridge this coordinate gap.


To represent the geometric and textural information of the generated object ${\mathcal{G}}_i^{\mathrm{gen}}$ in 2D space, we render multi-view images from $V$ predefined viewpoints:
\begin{equation}
\{\mathbf{I}_{i,v}^{\mathrm{gen}}\}_{v=1}^{V} = \Pi_{\mathrm{render}}\!\left(\mathcal{G}_i^{\mathrm{gen}};\, \{\mathbf{K}_v, \mathbf{E}_v^{\mathrm{obj}}\}_{v=1}^{V}\right),
\end{equation}
where \(\mathbf{K}_v \in \mathbb{R}^{3\times3}\) and \(\mathbf{E}_v^{\mathrm{obj}} = [\mathbf{R}_v^{\mathrm{obj}} \mid \mathbf{t}_v^{\mathrm{obj}}]\) denote the intrinsic matrix and extrinsic parameters for the \(v\)-th view. 

As for the input structure, the vanilla VGGT model designates the first input image in the sequence as the reference frame for reconstruction. 
To adapt to this mechanism, we regard the perspective crop $\mathbf{I}^\text{crop}_i$ as the target image for alignment and set it as the first input image.
Then, we concatenate it with multi-view renderings $\{\mathbf{I}_{i,v}^{\mathrm{gen}}\}_{v=1}^{V}$ as input to Alignment-VGGT.
Moreover, we also provide known camera parameters to avoid the intrinsic–extrinsic ambiguity when jointly estimating both parameters from images alone. Specifically, for each view (target crop or a multi-view rendered image), we encode its intrinsic matrix $\mathbf{K} \in \mathbb{R}^{3\times3}$ and extrinsic parameters $\mathbf{E}=[\mathbf{R}\mid\mathbf{t}]$ via two separate linear projection layers, and then add this camera embedding to the corresponding camera token for that view as input to the transformer layers.
This design explicitly disentangles camera geometry from visual features, enabling Alignment-VGGT to accurately resolve the cross-coordinate mapping even when intrinsics and scales differ between the local object frame and the world panorama frame.

Regarding the output head, the camera head in vanilla VGGT predicts the rotation and translation components of the extrinsics, omitting the scale transformation.
We resolve this by augmenting Alignment-VGGT with a scale head, enabling the model to output a complete set of camera extrinsics for the local coordinates and the anisotropic scale factors for mapping to world coordinates.


The forward pass of Alignment-VGGT is formulated as:
\begin{equation}
\{\hat{\mathcal{E}}, \hat{\mathbf{S}}\} = \mathcal{F}_{\mathrm{a-vggt}}\left(\mathbf{I}_i^{\mathrm{crop}}, \{\mathbf{I}_{i,v}^{\mathrm{gen}}\}_{v=1}^{V}, \{\mathbf{K}_v\}_{v=0}^{V}, \{\mathbf{E}_v^{\mathrm{obj}}\}_{v=1}^{V}\right).
\end{equation}

The network inputs include (i) the target crop $\mathbf{I}_i^{\mathrm{crop}}$ and its intrinsics $\mathbf{K}_0$, (ii) multi-view renderings $\{\mathbf{I}_{i,v}^{\mathrm{gen}}\}_{v=1}^{V}$ with their known intrinsics $\{\mathbf{K}_v\}_{v=1}^{V}$ and extrinsics $\{\mathbf{E}_v^{\mathrm{obj}}\}_{v=1}^{V}$ in the local frame. Note that $\mathbf{E}_0^{\mathrm{obj}}$ is unknown and not provided as input. 
The network outputs predicted poses for all views and anisotropic scale:
\begin{equation}
\hat{\mathcal{E}}=\left\{ \hat{\mathbf{E}}_v = [\hat{\mathbf{R}}_v^{\mathrm{obj}} \mid \hat{\mathbf{t}}_v^{\mathrm{obj}}] \right\}_{v=0}^V, \hat{\mathbf{S}} = \mathrm{diag}(\hat{s}_x,\hat{s}_y,\hat{s}_z),
\end{equation}
which are defined in Aignment-VGGT coordinate system. 

Then, we infer the unknown local extrinsics by relative pose chaining. 
First, we compute the coordinate-invariant relative transformation from view 1 to view 0 in the Aignment-VGGT coordinate system, and then apply it to the known extrinsics in the object's local coordinate system:
\begin{equation}
\mathbf{E}^{\mathrm{obj}}_0 = \Delta\mathbf{E}_{1\to 0} \mathbf{E}^{\mathrm{obj}}_1,
\end{equation}
where $\Delta\mathbf{E}_{1\to 0} = \hat{\mathbf{E}}_0 \hat{\mathbf{E}}_1^{-1}$.

Given $\mathbf{E}^{\mathrm{obj}}_0$ in the local frame and $\mathbf{E}_i^{\mathrm{crop}}=[\mathbf{R}_i^\text{w}\mid\mathbf{t}_i^\text{w}]$ in the world frame, we compose the non-rigid transformation incorporating the predicted anisotropic scale:
\begin{equation}
\mathbf{T}_i =
\begin{bmatrix}
\mathbf{R}_i^{\mathrm{w}} & \mathbf{t}_i^{\mathrm{w}} \\
\mathbf{0}^\top & 1
\end{bmatrix}
\begin{bmatrix}
(\mathbf{R}^{\mathrm{obj}}_0)^\top & -(\mathbf{R}^{\mathrm{obj}}_0)^\top \mathbf{t}^{\mathrm{obj}}_0 \\
\mathbf{0}^\top & 1
\end{bmatrix}
\begin{bmatrix}
\hat{\mathbf{S}} & \mathbf{0} \\
\mathbf{0}^\top & 1
\end{bmatrix}.
\label{eq:final_T}
\end{equation}

Finally, we apply the transformation $\mathbf{T}_i$ to convert the generated object from the local coordinate system to the world coordinate system:
\begin{equation}
\mathcal{G}_i^{\mathrm{w}} = \{ \mathbf{T}_i\,\mathbf{p} \mid \mathbf{p}\in\mathcal{G}_i^{\mathrm{gen}}\},
\end{equation}
where each point $\mathbf{p} = [x^\text{obj}, y^\text{obj}, z^\text{obj}, 1]^\top$ belongs to the 3D point set $\mathcal{P}_i^\mathrm{gen}$ of $\mathcal{G}_i^{\mathrm{gen}}$. For mesh representations, we simply transform vertex positions. For 3D Gaussian Splatting representations, in addition to transforming Gaussian centers, we also transform their covariance matrices by applying the rotation and scale components of $\mathbf{T}_i$ to ensure proper orientation and shape in the world frame.

\subsubsection{Loss Functions}
\label{sec:loss}
After obtaining the transformed object $\mathcal{G}_i^{\mathrm{w}}$ via the predicted transformation, a key challenge arises in training the predictor: directly supervising with ground-truth (GT) mesh poses is infeasible due to inevitable shape discrepancies between the generated object $\mathcal{G}_i^{\mathrm{gen}}$ and the GT mesh $\mathcal{G}_i^{\mathrm{GT}}$. Even if GT pose annotations were available, they would correspond to the GT geometry rather than the generated geometry, leading to misaligned supervision signals.

To address this issue, we adopt a pseudo-geometry supervision scheme that distills transformation parameters from slow but reliable offline optimizers. For each generated object, we run an offline differentiable optimization to fit rotation $\mathbf{R}\in\mathrm{SO}(3)$, translation $\mathbf{t}\in\mathbb{R}^3$, and anisotropic scale $\mathbf{S}=\mathrm{diag}(s_x,s_y,s_z)$. The resulting parameters $(\mathbf{R}^\star,\mathbf{t}^\star,\mathbf{S}^\star)$ define a transformation from the generated object to its GT mesh. 

\noindent\textbf{Supervision with GT meshes.}
When GT 3D meshes are available, we optimize the transform using a bidirectional Chamfer loss. Let $\mathcal{P}^{\mathrm{w}}$ denote the set of points sampled from the generated object transformed from the $i$-th object, and $\mathcal{P}$ be points from the GT mesh. We define:

{\footnotesize
\begin{equation}
\mathcal{L}_{\mathrm{CD}}^{\mathrm{bi}}(\mathcal{P}^{\mathrm{w}},\mathcal{P}) =
\frac{1}{|\mathcal{P}^{\mathrm{w}}|}\sum_{\hat{\mathbf{p}}\in\mathcal{P}^{\mathrm{w}}}\min_{\mathbf{p}\in\mathcal{P}}\|\hat{\mathbf{p}}-\mathbf{p}\|_2^2
+
\frac{1}{|\mathcal{P}|}\sum_{\mathbf{p}\in\mathcal{P}}\min_{\hat{\mathbf{p}}\in\mathcal{P}^{\mathrm{w}}}\|\mathbf{p}-\hat{\mathbf{p}}\|_2^2,
\end{equation}
}
the offline optimization minimizes:
\begin{equation}
(\mathbf{R}^\star,\mathbf{t}^\star,\mathbf{S}^\star)
=\arg\min_{\mathbf{R},\mathbf{t},\mathbf{S}}\;\mathcal{L}_{\mathrm{CD}}^{\mathrm{bi}}(\mathcal{P}^{\mathrm{w}},\mathcal{P}).
\end{equation}

\noindent\textbf{Supervision with monocular RGBD.}
When GT meshes are unavailable, we back-project GT depth into a partial point cloud $\mathcal{P}$ and use a single-directional Chamfer loss:
\begin{equation}
\mathcal{L}_{\mathrm{CD}}^{\mathrm{si}}(\mathcal{P}^{\mathrm{w}},\mathcal{P}) =
\frac{1}{|\mathcal{P}|}\sum_{\mathbf{p}\in\mathcal{P}}\min_{\hat{\mathbf{p}}\in\mathcal{P}^{\mathrm{w}}}\|\mathbf{p}-\hat{\mathbf{p}}\|_2^2.
\end{equation}

To mitigate inaccuracies from asymmetric point-to-surface matching, we augment $\mathcal{L}_{\mathrm{CD}}^{\mathrm{si}}$ with a mask loss 
\begin{equation}
\mathcal{L}_{\mathrm{MASK}} = \|\mathbf{M}-\hat{\mathbf{M}}\|_2^2 + 1 - \mathrm{IoU}(\mathbf{M},\hat{\mathbf{M}}),
\end{equation}
where $\hat{\mathbf{M}}$ denotes the rendered mask and $\mathbf{M}$ denotes the instance mask. The optimization becomes:
\begin{equation}
(\mathbf{R}^\star,\mathbf{t}^\star,\mathbf{S}^\star)
=\arg\min_{\mathbf{R},\mathbf{t},\mathbf{S}}\;\big(\mathcal{L}_{\mathrm{CD}}^{\mathrm{si}}(\mathcal{P}^{\mathrm{w}},\mathcal{P})+\lambda_{\mathrm{MASK}}\mathcal{L}_{\mathrm{MASK}}\big).
\end{equation}

\noindent\textbf{Training objective.}
During training of the Object-World Transformation Predictor, given transformation parameters \((\mathbf{R}^\star,\mathbf{t}^\star,\mathbf{S}^\star)\) and predictions \((\hat{\mathbf{R}},\hat{\mathbf{t}},\hat{\mathbf{S}})\), we regress parameters with element-wise L1 losses. For rotation, we convert to unit quaternions and, after normalization and sign alignment, apply element-wise L1. Let \(\hat{\mathbf{q}}=\mathrm{unit}(\mathrm{Q}(\hat{\mathbf{R}}))\) and \(\mathbf{q}^\star=\mathrm{unit}(\mathrm{Q}(\mathbf{R}^\star))\). The PGD loss is
\begin{equation}
\mathcal{L}_{\mathrm{PGD}}=\|\tilde{\mathbf{q}}-\mathbf{q}^\star\|_1+\|\hat{\mathbf{t}}-\mathbf{t}^\star\|_1+\|\mathrm{diag}(\hat{\mathbf{S}})-\mathrm{diag}(\mathbf{S}^\star)\|_1.
\end{equation}

We also include the mask loss to enforce silhouette consistency between rendered mask of the transformed object and the input instance mask. The total training objective of Object-World Transformation Predictor is
\begin{equation}
\mathcal{L}=\lambda_{\mathrm{CD}}\mathcal{L}_{\mathrm{CD}}+\lambda_{\mathrm{PGD}}\mathcal{L}_{\mathrm{PGD}}+\lambda_{\mathrm{MASK}}\mathcal{L}_{\mathrm{MASK}}.
\end{equation}

These serve as supervisory targets for the Object-World Transformation Predictor, replacing expensive per-instance optimization at inference with a single feed-forward pass.


\subsection{Background Reconstruction and Scene Fusion}
\label{sec:background}
We merge all instance masks and apply an inpainting model (LaMa \cite{suvorov2022lama} or DiT360 \cite{feng2025dit360}) on the panoramic image to obtain a clean background panorama $\mathbf{I}^{\mathrm{bg}}$. A feed-forward Gaussian reconstruction network, following Flash3D \cite{szymanowicz2025flash3d}, predicts background depth with Depth-Anywhere \cite{wang2024depthanywhere} and generates a background Gaussian set $\mathcal{G}_{\mathrm{bg}}$. Finally, we place all aligned instances $\{\mathcal{G}^\text{w}_i\}$ and fuse them with $\mathcal{G}_{\mathrm{bg}}$ in the world frame, producing the complete 3D scene \(\{\mathcal{G}_i^\text{w}\} \cup \mathcal{G}_{\mathrm{bg}}\).

\begin{figure}[t]
    \centering
    \includegraphics[width=0.9\columnwidth]{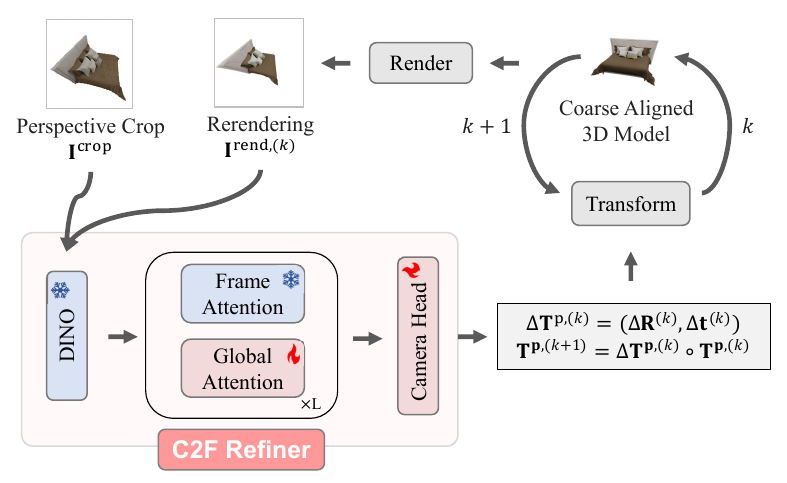}
    
    \vspace{-0.2cm}
    \caption{
        Illustration of the proposed Coarse-to-Fine (C2F) alignment mechanism.
    }
    \vspace{-0.3cm}
    \label{fig:modules}
\end{figure}

\section{Iterative Extension: Pano3DComposer-C2F}
\label{sec:refine}
The initial placement produced by the Object-World Transformation Predictor may be imperfect when applied to unseen domains, due to distribution shift between training and testing data. To address this, we introduce a Coarse-to-Fine (C2F) alignment mechanism to extend Pano3DComposer to an iterative version called Pano3DComposer-C2F, which progressively refines the object position by more steps. 

Compared with the Pano3DComposer, we additionally introduce a C2F Refiner module based on Alignment-VGGT with feedback of current rendering result.
Given a coarsely aligned object at pose $\mathbf{T}^{\mathrm{p},(k)}=(\mathbf{R}^{(k)},\mathbf{t}^{(k)})$ where $\mathbf{R}^{(k)}\in\mathrm{SO}(3)$ and $\mathbf{t}^{(k)}\in\mathbb{R}^3$, we render it using the camera parameters of the perspective crop to obtain the current rendering $\mathbf{I}^{\mathrm{rend},(k)}$. The refiner takes as input the concatenation of the current rendering $\mathbf{I}^{\mathrm{rend},(k)}$ and the target object crop $\mathbf{I}^{\mathrm{crop}}$ from the panorama, and estimates a relative pose update $\Delta\mathbf{T}^{\mathrm{p},(k)}=(\Delta\mathbf{R}^{(k)},\Delta\mathbf{t}^{(k)})$ while keeping the scale fixed to avoid shape distortion:
\begin{equation}
\Delta\mathbf{T}^{\mathrm{p},(k)} = \mathcal{F}_{\mathrm{refine}}\big(\mathbf{I}^{\mathrm{rend},(k)}, \mathbf{I}^{\mathrm{crop}}\big).
\end{equation}

The pose is then updated iteratively via composition:
\begin{equation}
\mathbf{T}^{\mathrm{p},(k+1)} = \Delta\mathbf{T}^{\mathrm{p},(k)} \circ \mathbf{T}^{\mathrm{p},(k)}, \quad k=0,1,\dots,K_{\max}-1,
\end{equation}
where $\mathbf{T}^{\mathrm{p},(0)}$ is the initial coarse pose from $\mathcal{F}_{\mathrm{pred}}$.


During training, we supervise the refiner with the same pseudo-geometry distillation loss in Section~\ref{sec:loss}. At inference time, we back-project depth estimated from the panorama into a point cloud $\mathcal{P}_{\mathrm{pseudo}}$ and monitor alignment quality via Chamfer distance $\mathcal{L}_{\mathrm{CD}}^{(k)}$ between the transformed object points and $\mathcal{P}_{\mathrm{pseudo}}$. The iteration terminates when the improvement falls below a threshold $\tau$ or the maximum iteration $K_{\max}$ is reached:
\begin{equation}
\mathcal{L}_{\mathrm{CD}}^{(k)} - \mathcal{L}_{\mathrm{CD}}^{(k+1)} < \tau \quad \text{or} \quad k+1=K_{\max}.
\end{equation}

This yields robust, feed-forward fine alignment without gradient-based optimization at test time, progressively correcting pose errors through rendering feedback.

\begin{table*}[t]
\centering
\caption{Comparison of major and alignment results on the 3D-FRONT test set. 
The best performance for each metric is highlighted in \textbf{bold}. OPT represents differentiable optimization-based alignment, and ICP denotes Iterative Closest Point alignment. ``Pseudo Geometry'' serves as a reference upper bound obtained via offline differentiable optimization of the transformation parameters introduced in Sec.~\ref{sec:loss}. Training resources are reported in 4090 GPU days. Inference time is tested on one 4090 GPU.}
\label{tab:main_results}
\resizebox{0.9\textwidth}{!}{
\begin{tabular}{lccccccc}
\hline
Method & CD-S$\downarrow$ & CD-O$\downarrow$ & F-Score-S$\uparrow$ & F-Score-O$\uparrow$ & IoU-B$\uparrow$ & Training Resources & Inference Time (s) \\
\hline
OPT & 0.1059 & 0.1128 & 0.5535 & 0.5640 & 0.4010 & -- & 120 \\
ICP \cite{small_gicp} & 0.2483 & 0.2305 & 0.4524 & 0.4896 & 0.2830 & -- & \textbf{1} \\
\hline
DeepPanoContext \cite{zhang2021deeppanocontext} & 0.7851 & 0.1657 & 0.3101 & 0.3822 & 0.0021 & -- & 14 \\
SceneGen \cite{meng2025scenegen} & 0.1765 & 0.0914 & 0.4575 & 0.4827 & 0.1124 & 56 GPU days & 63 \\
Pano3DComposer (Ours) & 0.0787 & 0.0765 & 0.6923 & 0.6926 & 0.5679 & \textbf{2 GPU days} & 20 \\
Pano3DComposer-C2F (Ours) & \textbf{0.0784} & \textbf{0.0762} & \textbf{0.6930} & \textbf{0.6937} & \textbf{0.5699} & 4 GPU days & 24 \\
\hline
\textit{Pseudo Geometry} & 0.0119 & 0.0119 & 0.8695 & 0.8781 & 0.8141 & -- & -- \\
\hline
\end{tabular}
}
\end{table*}

\begin{figure*}[t]
    \centering
    \includegraphics[width=0.9\textwidth]{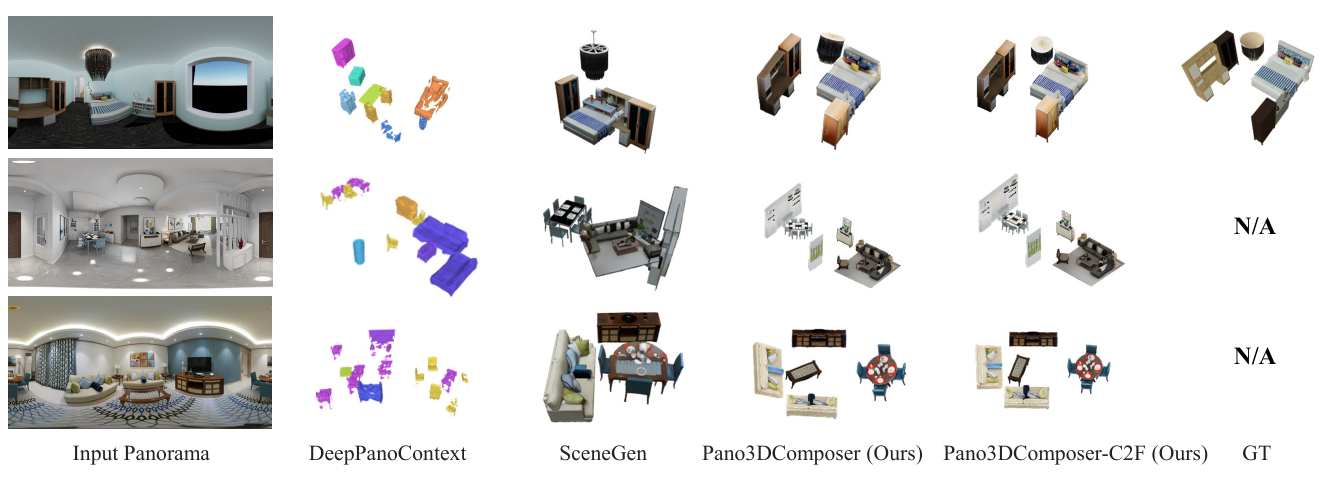}
    \vspace{-0.2cm}
    \caption{
        Visualization of panorama-to-3D scene composition results without background. Row 1: 3D-FRONT test set; Row 2: Structured3D test set; Row 3: real-world panoramas.
    }
    \vspace{-0.3cm}
    \label{fig:panorama_results}
\end{figure*}

\begin{figure*}[t]
    \centering
    \includegraphics[width=0.8\textwidth]{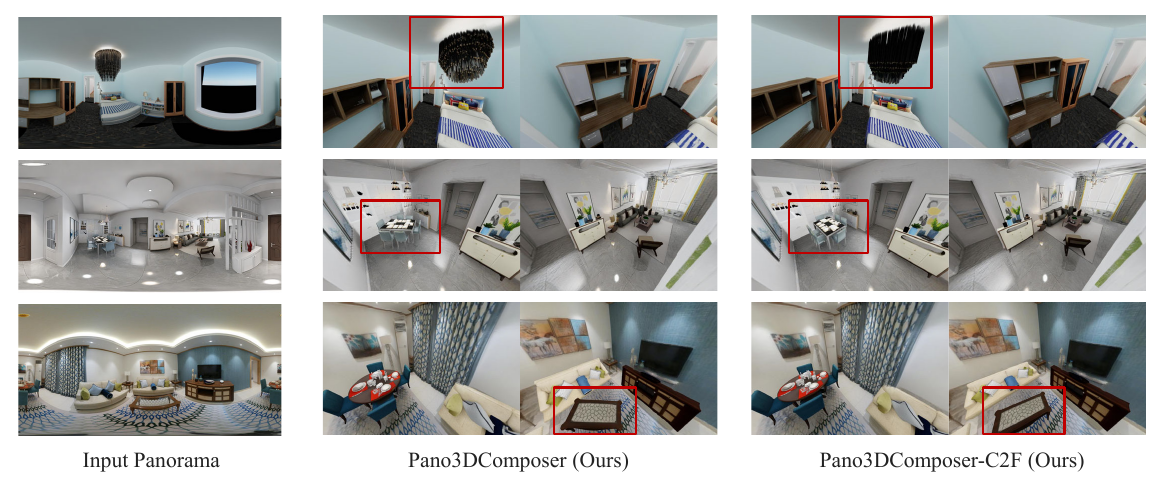}
    \vspace{-0.2cm}
    \caption{
        Visualization of panorama-to-3D scene composition results with background. The figure presents multi-view renderings of composed 3D scenes generated by our method. Row 1: 3D-FRONT test set; Row 2: Structured3D test set; Row 3: real-world panoramas.
    }
    \vspace{-0.3cm}
    \label{fig:panorama_results_mv}
\end{figure*}

\begin{figure}[t]
    \centering
    \includegraphics[width=\linewidth]{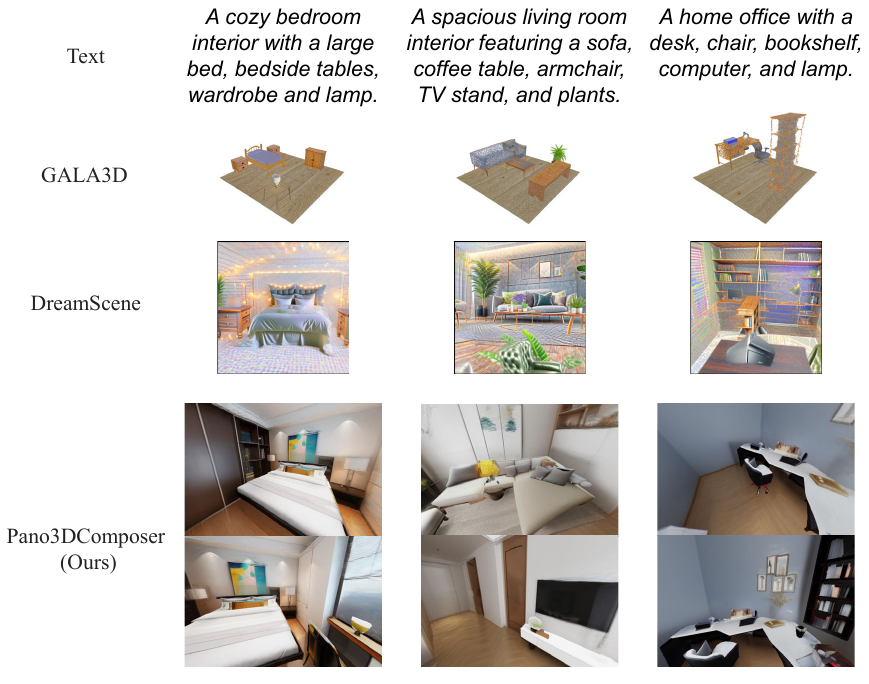}
    \vspace{-0.5cm}
    \caption{
        Visualization of Text-to-3D scene generation results.
    }
    \vspace{-0.5cm}
    \label{fig:text_results}
\end{figure}

\section{Experiments}
We conduct comprehensive experiments to validate the effectiveness of Pano3DComposer. Additional implementation details, qualitative results, visualizations, and ablation studies are provided in the supplementary material.

\subsection{Experimental Setup}
\noindent\textbf{Datasets.}
We train and evaluate our model on two large-scale synthetic indoor datasets: 3D-FRONT \cite{fu20213dfront} and Structured3D \cite{zheng2020structured3d}. 
For 3D-FRONT, we render equirectangular panoramas and corresponding depth maps for each room using Blender, and utilize the ground-truth 3D meshes to generate pseudo-geometry supervision via bidirectional Chamfer distance optimization (Sec.~\ref{sec:loss}). 
Since panoramas rendered from 3D-FRONT lack photorealism, we augment our training set with Structured3D, which provides photorealistic panoramas but only monocular depth without ground-truth meshes. For Structured3D, we employ single-directional Chamfer distance with mask regularization to derive pseudo-geometry supervision (Sec.~\ref{sec:loss}). 
In total, we collect approximately 30{,}000 rooms. with 1{,}200 held out for testing. Additionally, we evaluate our method on a collection of real-world panoramic images to assess generalization capability.

\noindent\textbf{Evaluation metric.}
We evaluate the shape and layout accuracy using scene-level Chamfer Distance (CD-S) and F-Score (F-Score-S), object-level Chamfer Distance (CD-O) and F-Score (F-Score-O), and volumetric IoU of object bounding boxes (IoU-B).

\noindent\textbf{Compared methods for panorama-to-3D scene composition.}
To comprehensively evaluate Pano3DComposer, which reconstructs a 3D scene from a single equirectangular panorama, we compare against DeepPanoContext \cite{zhang2021deeppanocontext}, a method specifically designed for panoramic inputs, and SceneGen \cite{meng2025scenegen}, a state-of-the-art feed-forward multi-instance generation method. Since SceneGen is designed exclusively for perspective images and cannot directly handle equirectangular panoramas due to distortion and non-uniform sampling, we fine-tune it on our panoramic 3D-FRONT dataset to enable a fair comparison with representative feed-forward approaches.

We also compare against two classical pose estimation baselines: Iterative Closest Point (ICP) \cite{besl1992icp} and differentiable optimization (OPT). For ICP, we use the implementation from the \textit{small\_gicp} library \cite{small_gicp} to align normalized point clouds extracted from the generated object and the reference image, followed by scaling with the estimated scale factor. For differentiable optimization, we optimize rotation, translation, and scale parameters to align the object with the reference RGB image and its corresponding depth prediction using a combination of photometric and geometric losses, as done in REPARO \cite{han2025reparo}. 

\noindent\textbf{Compared methods for text-to-3D scene generation.}
We further evaluate Pano3DComposer in the Text-to-3D Scene Generation setting, where our pipeline first synthesizes a panoramic image from text using Diffusion360~\cite{feng2023diffusion360}, and subsequently composes the corresponding 3D scene conditioned on the generated panorama. For this task, we include representative text-to-3D scene methods GALA3D \cite{Zhou2024GALA3D} and DreamScene \cite{li2024dreamscene}.

\subsection{Implementation Details}

For the feed-forward Gaussian background model, we follow the pipeline of Flash3D \cite{szymanowicz2025flash3d}, but replace its depth estimator with Depth-Anywhere \cite{wang2024depthanywhere} for improved monocular depth prediction on panoramas. For training the Object-World Transformation Predictor, we freeze the DINOv2 \cite{oquab2023dinov2} backbone and frame attention layers of VGGT \cite{wang2025vggt}. Mini-batch size is set as 1. Training on a single RTX 4090 takes roughly two days for both the Object-World Transformation Predictor and the C2F Refiner. The learning rate is set to $1\times10^{-4}$ with a cosine decay schedule. Loss weights are set as $\lambda_{\mathrm{CD}}=0.1$, $\lambda_{\mathrm{PGD}}=1.0$, and $\lambda_{\mathrm{MASK}}=0.1$.
For the C2F refinement stage, we set the Chamfer distance threshold to $\tau=0.001$ and the maximum iteration to $K_{\max}=5$. 
For a fair comparison with the SceneGen baseline, we fine-tune from its released checkpoint on 3D-FRONT dataset using 8$\times$RTX 4090 GPUs for 7 days to adapt it to equirectangular panoramic inputs. 

\subsection{Panorama-to-3D Scene Composition}
\label{sec:pano23d}

Table~\ref{tab:main_results} presents quantitative comparisons on the 3D-FRONT test set. Our Pano3DComposer achieves the best performance across all metrics while requiring significantly fewer training resources (1×4090, 2d vs. 8×4090, 7d for SceneGen) and faster inference (20s vs. 63s per scene). Results in Table~\ref{tab:main_results} also show that our predictor significantly outperforms both ICP and differentiable optimization. ICP struggles with outliers and symmetries, often converging to local minima. Differentiable optimization is sensitive to inaccurate depth estimates and occlusions, leading to suboptimal alignment. Our feed-forward predictor, trained with pseudo-geometry supervision, robustly estimates object poses from generated 3D models, achieving the best alignment performance.

Figure~\ref{fig:panorama_results} shows qualitative comparisons in the panorama-to-3D task. DeepPanoContext fails to generate high-quality meshes due to limited supervised data. SceneGen struggles with panoramic distortion, resulting in incorrect spatial relationships. Our method mitigates distortion through perspective projection and produces scenes with consistent geometry and plausible spatial relationships.

Pano3DComposer-C2F further improves alignment with marginal additional computational cost (24s vs. 20s). Beyond improvements on synthetic test set, the C2F mechanism also exhibits good generalization to real-world panoramas, as shown in Figure~\ref{fig:panorama_results_mv}. As highlighted in the red boxes, C2F mechanism effectively corrects object positions through iterative refinement with rendering feedback. This validates the necessity and effectiveness of our iterative refinement mechanism for practical applications, enabling robust generalization to unseen data distributions without requiring expensive per-scene optimization. Table~\ref{tab:runtime} provides a detailed breakdown of computational costs. These results demonstrate that our feed-forward transformation predictor and modular design enable accurate compositional 3D scene generation without sacrificing efficiency.

\subsection{Text-to-3D Scene Generation}
\label{sec:text23d}

As shown in Figure~\ref{fig:text_results}, both GALA3D and DreamScene rely on Score Distillation Sampling (SDS) to optimize object appearance, which requires lengthy per-scene optimization (typically 30-60 minutes per object) and often leads to oversaturated colors and unrealistic textures. Moreover, these methods depend on LLM-generated layouts to determine spatial arrangements, which frequently violate physical constraints and common spatial relationships. For instance, objects may float in mid-air, penetrate each other, or be placed at physically implausible positions.

In contrast, Pano3DComposer leverages state-of-the-art image-to-3D object generators to produce high-fidelity textured meshes, and derives spatial layouts directly from the synthesized panoramic image through our feed-forward transformation predictor. This design naturally respects the spatial context encoded in the panorama, ensuring physically plausible object arrangements. As a result, our method generates scenes with more realistic textures, accurate spatial relationships, and significantly improved efficiency.

\begin{table}[t]
\caption{
Runtime analysis of different processing stages. 
}
\centering
\resizebox{0.7\linewidth}{!}{
\begin{tabular}{l c}
\hline
Stage & Time (s) \\ \hline
Background Inpainting & 0.02 \\
Background GS Reconstruction & 0.16 \\ \hline
Object Generation (per object) & $\sim$4 \\
Object Alignment (per object) & 0.36 \\
Object Refinement (per step) & 0.18 \\ \hline
\end{tabular}
}
\label{tab:runtime}
\end{table}


\begin{table}[]
\caption{
Ablation study on loss functions and training strategies. 
}
\resizebox{\linewidth}{!}{
\begin{tabular}{lccccc}
\hline
Method    & CD-S $\downarrow$ & CD-O $\downarrow$ & F-Score-S $\uparrow$ & F-Score-O $\uparrow$ & IoU-B $\uparrow$ \\ \hline
Only $\mathcal{L}_{\mathrm{CD}}$       & 0.8688                     & 0.9027                     & 0.1980                  & 0.1888                  & 0.0906                  \\
+ $\mathcal{L}_{\mathrm{PGD}}$      & 0.1266                     & 0.1219                     & 0.5675                  & 0.5670                  & 0.4670                  \\
+ $\mathcal{L}_{\mathrm{MASK}}$ & \textbf{0.1120}            & \textbf{0.1063}            & \textbf{0.5788}                  & \textbf{0.5850}                  & \textbf{0.4818}                  \\ \hline
w/o Cam info       & 0.1850                     & 0.1705                     & 0.4673                  & 0.4691                  & 0.3830                  \\ \hline
\end{tabular}
}
\label{tab:ablation}
\end{table}

\subsection{Ablation Study}

Table~\ref{tab:ablation} compares the contributions of different losses and training strategies. Training with only the Chamfer loss $\mathcal{L}_{\mathrm{CD}}$ yields poor alignment quality, as the model fails to learn accurate pose regression without direct supervision on transformation parameters, introducing the pseudo-geometry distillation loss $\mathcal{L}_{\mathrm{PGD}}$ substantially improves all metrics. Further adding the mask regularization $\mathcal{L}_{\mathrm{MASK}}$ brings additional gains, indicating that silhouette consistency is essential for spatial coherence. Excluding known camera intrinsics and extrinsics as input leads to a noticeable performance drop, further validating the importance of incorporating camera priors.

\section{Conclusion}


We presented Pano3DComposer, an efficient framework for compositional 3D scene generation from a single panorama. By learning a feed-forward Object-World Transformation Predictor with pseudo-geometry supervision, and applying a C2F alignment mechanism, our approach outperforms state-of-the-art methods across all metrics on 3D-FRONT dataset, and generalizes robustly to real-world panoramas through the C2F refinement mechanism. Our method generates high-fidelity 3D scenes in approximately 20 seconds per scene, making it practical for real-time applications in VR/AR and digital content creation.

{
    \small
    \bibliographystyle{ieeenat_fullname}
    \bibliography{main}
}

\input{supp}

\end{document}

%% file: supp.tex
\clearpage
\setcounter{page}{1}

\maketitlesupplementary

\section*{Overview}
This supplementary material provides additional details, including dataset descriptions, implementation details, ablation studies, qualitative results, failure cases, and limitations to complement the main paper. We also provide supplementary videos showcasing qualitative results and rendered 3D scenes, which further demonstrate the effectiveness of our method.

\section{Datasets}
Our experiments involve panorama-to-3D scene composition on synthetic benchmarks and real-world panoramas. Below we summarize the synthetic datasets used for training and quantitative evaluation. 3D-FRONT \cite{fu20213dfront} is a professionally designed dataset comprising high-quality textured furniture models arranged in realistic room layouts. Structured3D \cite{zheng2020structured3d} is a photo-realistic synthetic dataset featuring rendered images under diverse lighting and furniture configurations, accompanied by rich annotations (semantics, albedo, depth, normals, and layout) but does not release object meshes. For real-world in-the-wild panoramas used in qualitative evaluation, we collect images from public online sources and ensure they are only used for non-commercial research visualization.

\begin{figure}[b]
    \centering
    \includegraphics[width=\columnwidth]{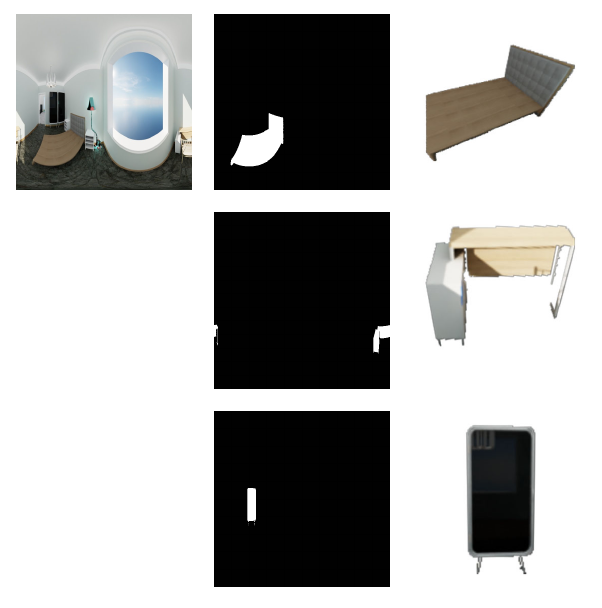}
    \caption{Example inputs of SceneGen.}
    \label{fig:erp_example}
\end{figure}

\section{More Implementation Details}

\noindent\textbf{Fine-tune of SceneGen.} To adapt SceneGen \cite{meng2025scenegen} for equirectangular panoramic (ERP) inputs, we follow its official data preprocessing pipeline and extend it to handle the equirectangular panoramas rendered from the 3D-FRONT dataset. A representative example of the processed panoramic input, including the panorama, instance masks, and object crops, is shown in Fig.~\ref{fig:erp_example}. Our model is initialized from the official SceneGen pretrained checkpoint. We fine-tune the model using a global batch size of $8$ and an initial learning rate of $1\times 10^{-5}$ with AdamW optimizer. The model is trained for $7$ days on a single NVIDIA RTX~4090 GPU under mixed-precision (BF16) training.

\noindent\textbf{Inference.} In our experiments, the input equirectangular panoramas are at a resolution of $512 \times 1024$. For evaluation on 3D-FRONT and Structured3D, we directly use the ground-truth instance segmentation annotations provided by each dataset, in the same way as SceneGen \cite{meng2025scenegen}. 
For real-world in-the-wild data, we manually obtain instance masks using the 2D foundation model SAM \cite{kirillov2023sam}. To develop a fully automated pipeline, one may integrate open-vocabulary recognition models (e.g., RAM \cite{zhang2024ram}, various visual language models (VLMs)) with detection/segmentation models capable of grounding (e.g., GroundingDINO \cite{liu2024groundingdino}, SAM \cite{kirillov2023sam}, Grounded-SAM \cite{ren2024groundedsam}) to identify, localize, and segment objects directly on ERP panoramas.
In the Object-World Transformation Predictor, we render $4$ multi-view images for each object. We uniformly sample four horizontal viewing directions at azimuth angles $\{0^{\circ}, 90^{\circ}, 180^{\circ}, 270^{\circ}\}$, and apply a fixed $20^{\circ}$ downward pitch. All renderings use a resolution of $518 \times 518$. These rendered views provide appearance-conditioned geometric cues that significantly stabilize the relative pose estimation stage.

\begin{table}[]
\caption{Ablation of fine-tuning strategies. ``–D'', ``–D-F'', and ``–D-F-G'' indicate progressively freezing DINO, frame, and global attention modules.}

\resizebox{\linewidth}{!}{
\begin{tabular}{lccccc}
\hline
\textbf{Method} & \textbf{CD-S $\downarrow$} & \textbf{CD-O $\downarrow$} & \textbf{F-Score-S $\uparrow$} & \textbf{F-Score-O $\uparrow$} & \textbf{IoU-B $\uparrow$} \\ \hline
Full                 & 0.1883 & 0.1946 & 0.4992 & 0.4907 & 0.3855 \\
–D             & 0.1236 & 0.1177 & 0.5565 & 0.5550 & 0.4360 \\
–D-F            & \textbf{0.0787} & \textbf{0.0765} & \textbf{0.6923} & \textbf{0.6926} & \textbf{0.5679} \\
–D-F-G   & 0.1120 & 0.1063 & 0.5788 & 0.5850 & 0.4818 \\ \hline
\end{tabular}
}
\label{tab:ablation_ft}
\end{table}

\begin{figure*}[t]
    \centering
    \includegraphics[width=\textwidth]{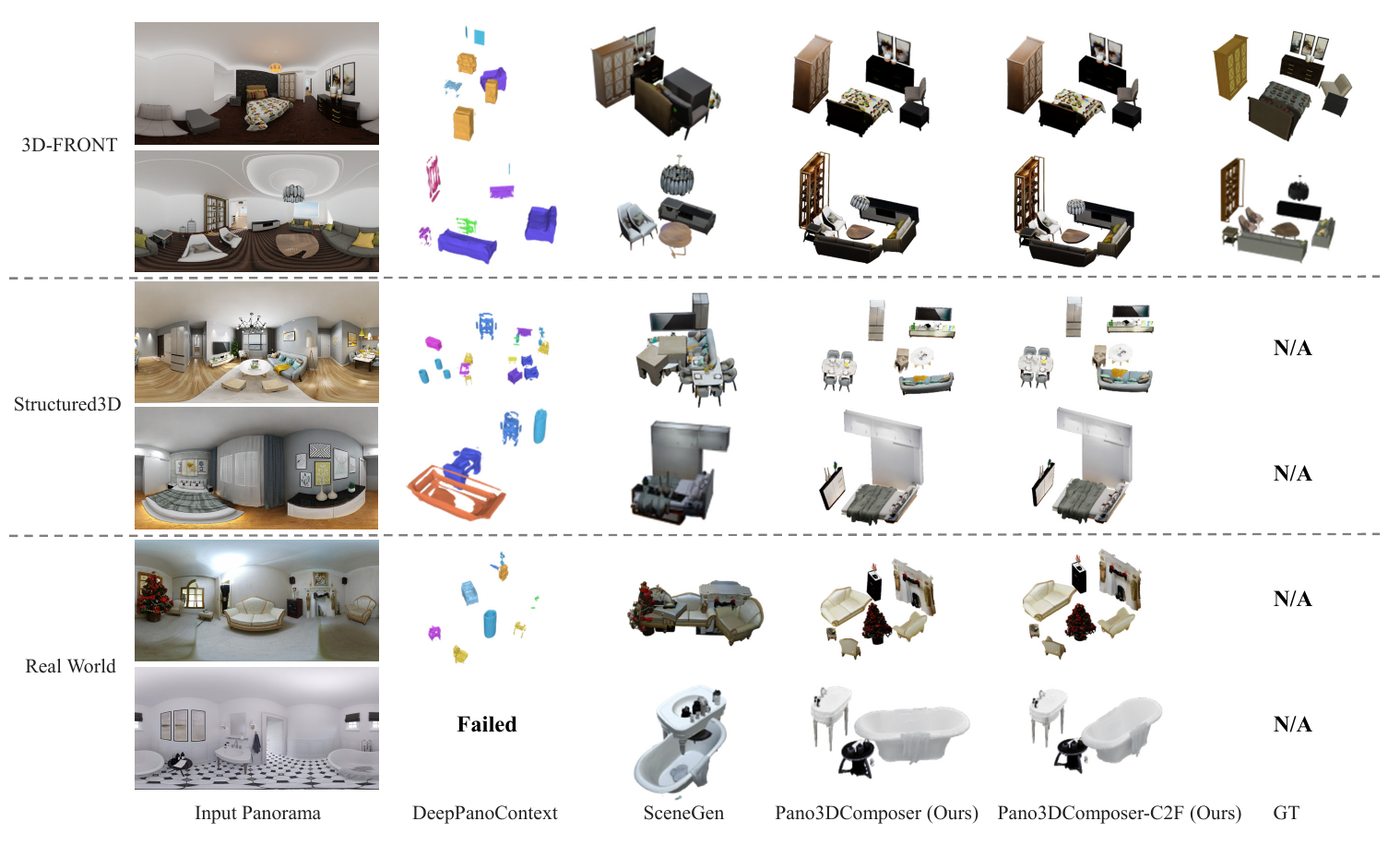}
    \caption{
        Visualization of panorama-to-3D scene composition results without background.
    }
    \label{fig:supp_result}
\end{figure*}

\section{More Experiments}
\subsection{Ablation of Trainable VGGT Modules.}
We compare different fine-tuning strategies by freezing specific modules of VGGT \cite{wang2025vggt} (Table \ref{tab:ablation_ft}).
``Full'' denotes full fine-tuning. ``–D'' freezes the DINO backbone; ``–D-F'' further freezes the frame attention layers; and ``–D-F-G'' also freezes the global attention layers. We find that keeping the global attention and camera/scale heads trainable (``–D-F'') yields the largest performance gains across all metrics.
\subsection{More Visual Comparisons}

Fig.~\ref{fig:supp_result} shows additional qualitative comparisons between our approach and baselines. To better illustrate the full-room generation capability beyond object synthesis, we provide more rendered videos in the supplementary attachment.

\begin{figure}
    \centering
    \includegraphics[width=\linewidth]{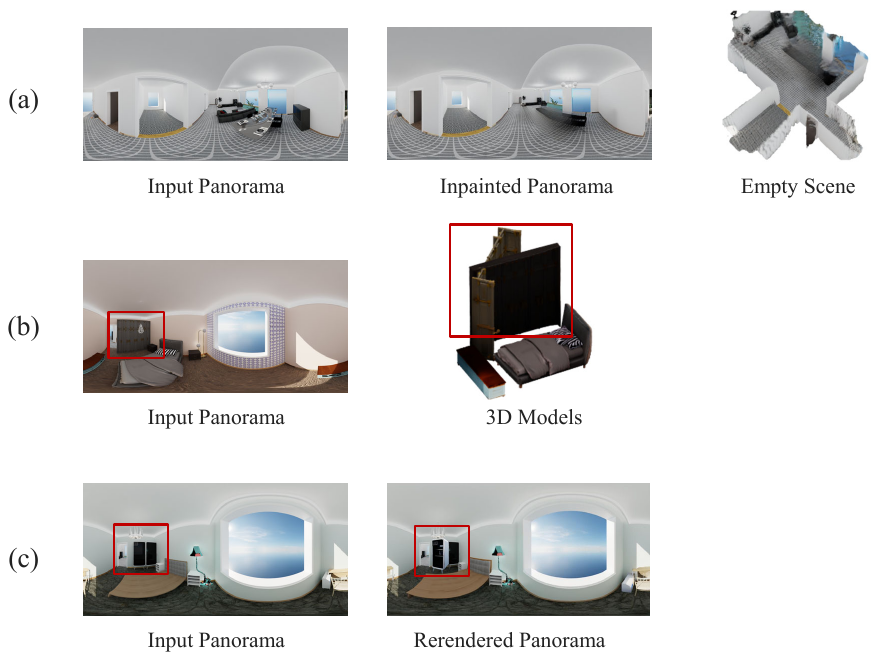}
    \caption{Failure cases. (a) Background inpainting and Flash3D-based monocular reconstruction failures. (b) Object generation failures. (c) Alignment failures.}
    \label{fig:badcase}
\end{figure}

\section{Failure Cases}

When backgrounds exhibit complex geometry, clutter, or heavy occlusions, the inpainting network may fail to recover a clean room structure. This can lead to visible artifacts or incorrect structural completions. In addition, the Flash3D-based \cite{szymanowicz2025flash3d} monocular reconstruction is affected by the quality of depth estimation; inaccurate depth may lead to distorted backgrounds and other artifacts, as illustrated in Fig.~\ref{fig:badcase} (a). Moreover, since the input panorama is constrained to a resolution of $512 \times 1024$, the extracted object crops often have relatively low resolution. As a result, object generation models (e.g., TRELLIS \cite{xiang2025trellis}) may occasionally produce suboptimal outputs or even fail to generate plausible results (Fig.~\ref{fig:badcase} (b)).

When the generated 3D object differs drastically from the observed object in the input panorama (in terms of geometry, silhouette, or texture), or when objects in the panorama appear at very low resolution, the alignment network may fail to reliably estimate the relative pose, resulting in misaligned insertions (Fig.~\ref{fig:badcase} (c)).

\section{Limitations}
Our approach primarily targets indoor scenes. Very small items and highly articulated or multi-part objects can still exhibit residual misalignment. Highly glossy or transparent materials pose challenges for appearance modeling and silhouette consistency. Future work includes: (i) integrating physical awareness and multi-instance relation modeling, (ii) improving appearance and geometry prediction for transparent/specular objects, and (iii) scaling training data realism and diversity to further improve generalization.

%% file: main.bib
@String(CVPR= {IEEE Conf. Comput. Vis. Pattern Recog.})

@String(ICCV= {Int. Conf. Comput. Vis.})

@String(ECCV= {Eur. Conf. Comput. Vis.})

@String(NIPS= {Adv. Neural Inform. Process. Syst.})

@String(TOG= {ACM Trans. Graph.})

@String(CVPR  = {CVPR})

@String(ICCV  = {ICCV})

@String(ECCV  = {ECCV})

@String(NIPS  = {NeurIPS})

@String(TOG   = {ACM TOG})

@article{poole2022dreamfusion,
  title={Dreamfusion: Text-to-3d using 2d diffusion},
  author={Poole, Ben and Jain, Ajay and Barron, Jonathan T and Mildenhall, Ben},
  journal={arXiv},
  year={2022}
}

@inproceedings{lin2023magic3d,
  title={Magic3d: High-resolution text-to-3d content creation},
  author={Lin, Chen-Hsuan and Gao, Jun and Tang, Luming and Takikawa, Towaki and Zeng, Xiaohui and Huang, Xun and Kreis, Karsten and Fidler, Sanja and Liu, Ming-Yu and Lin, Tsung-Yi},
  booktitle={CVPR},
  year={2023}
}

@article{kerbl20233dgs,
  title={3D Gaussian splatting for real-time radiance field rendering.},
  author={Kerbl, Bernhard and Kopanas, Georgios and Leimk{\"u}hler, Thomas and Drettakis, George},
  journal={TOG},
  year={2023}
}

@inproceedings{yi2024gaussiandreamer,
  title={Gaussiandreamer: Fast generation from text to 3d gaussians by bridging 2d and 3d diffusion models},
  author={Yi, Taoran and Fang, Jiemin and Wang, Junjie and Wu, Guanjun and Xie, Lingxi and Zhang, Xiaopeng and Liu, Wenyu and Tian, Qi and Wang, Xinggang},
  booktitle={CVPR},

  year={2024}
}

@inproceedings{liang2024luciddreamer,
  title={Luciddreamer: Towards high-fidelity text-to-3d generation via interval score matching},
  author={Liang, Yixun and Yang, Xin and Lin, Jiantao and Li, Haodong and Xu, Xiaogang and Chen, Yingcong},
  booktitle={CVPR},
  year={2024}
}

@article{li2023gaussiandiffusion,
  title={Gaussiandiffusion: 3d gaussian splatting for denoising diffusion probabilistic models with structured noise},
  author={Li, Xinhai and Wang, Huaibin and Tseng, Kuo-Kun},
  journal={arXiv},
  year={2023}
}

@inproceedings{liu2023zero123,
  title={Zero-1-to-3: Zero-shot one image to 3d object},
  author={Liu, Ruoshi and Wu, Rundi and Van Hoorick, Basile and Tokmakov, Pavel and Zakharov, Sergey and Vondrick, Carl},
  booktitle={ICCV},

  year={2023}
}

@article{shi2023zero123++,
  title={Zero123++: a single image to consistent multi-view diffusion base model},
  author={Shi, Ruoxi and Chen, Hansheng and Zhang, Zhuoyang and Liu, Minghua and Xu, Chao and Wei, Xinyue and Chen, Linghao and Zeng, Chong and Su, Hao},
  journal={arXiv},
  year={2023}
}

@article{shi2023mvdream,
  title={Mvdream: Multi-view diffusion for 3d generation},
  author={Shi, Yichun and Wang, Peng and Ye, Jianglong and Long, Mai and Li, Kejie and Yang, Xiao},
  journal={arXiv},
  year={2023}
}

@article{liu2023syncdreamer,
  title={Syncdreamer: Generating multiview-consistent images from a single-view image},
  author={Liu, Yuan and Lin, Cheng and Zeng, Zijiao and Long, Xiaoxiao and Liu, Lingjie and Komura, Taku and Wang, Wenping},
  journal={arXiv},
  year={2023}
}

@article{hong2023lrm,
  title={Lrm: Large reconstruction model for single image to 3d},
  author={Hong, Yicong and Zhang, Kai and Gu, Jiuxiang and Bi, Sai and Zhou, Yang and Liu, Difan and Liu, Feng and Sunkavalli, Kalyan and Bui, Trung and Tan, Hao},
  journal={arXiv},
  year={2023}
}

@inproceedings{tang2024lgm,
  title={Lgm: Large multi-view gaussian model for high-resolution 3d content creation},
  author={Tang, Jiaxiang and Chen, Zhaoxi and Chen, Xiaokang and Wang, Tengfei and Zeng, Gang and Liu, Ziwei},
  booktitle={ECCV},

  year={2024}
}

@article{zhang2024clay,
  title={Clay: A controllable large-scale generative model for creating high-quality 3d assets},
  author={Zhang, Longwen and Wang, Ziyu and Zhang, Qixuan and Qiu, Qiwei and Pang, Anqi and Jiang, Haoran and Yang, Wei and Xu, Lan and Yu, Jingyi},
  journal={TOG},

  year={2024}
}

@inproceedings{xiang2025trellis,
  title={Structured 3d latents for scalable and versatile 3d generation},
  author={Xiang, Jianfeng and Lv, Zelong and Xu, Sicheng and Deng, Yu and Wang, Ruicheng and Zhang, Bowen and Chen, Dong and Tong, Xin and Yang, Jiaolong},
  booktitle={CVPR},
  year={2025}
}

@inproceedings{nie2020total3dunderstanding,
  title={Total3dunderstanding: Joint layout, object pose and mesh reconstruction for indoor scenes from a single image},
  author={Nie, Yinyu and Han, Xiaoguang and Guo, Shihui and Zheng, Yujian and Chang, Jian and Zhang, Jian Jun},
  booktitle={CVPR},
  year={2020}
}

@inproceedings{zhang2021im3d,
  title={Holistic 3d scene understanding from a single image with implicit representation},
  author={Zhang, Cheng and Cui, Zhaopeng and Zhang, Yinda and Zeng, Bing and Pollefeys, Marc and Liu, Shuaicheng},
  booktitle={CVPR},
  
  year={2021}
}

@inproceedings{liu2022instpifu,
  title={Towards high-fidelity single-view holistic reconstruction of indoor scenes},
  author={Liu, Haolin and Zheng, Yujian and Chen, Guanying and Cui, Shuguang and Han, Xiaoguang},
  booktitle={ECCV},
  
  year={2022}
}

@article{dahnert2024coherent,
  title={Coherent 3D scene diffusion from a single RGB image},
  author={Dahnert, Manuel and Dai, Angela and M{\"u}ller, Norman and Nie{\ss}ner, Matthias},
  journal={NIPS},
 
  year={2024}
}

@inproceedings{zhang2021deeppanocontext,
  title={Deeppanocontext: Panoramic 3d scene understanding with holistic scene context graph and relation-based optimization},
  author={Zhang, Cheng and Cui, Zhaopeng and Chen, Cai and Liu, Shuaicheng and Zeng, Bing and Bao, Hujun and Zhang, Yinda},
  booktitle={ICCV},
  year={2021}
}

@inproceedings{dong2024panocontext,
  title={PanoContext-Former: Panoramic total scene understanding with a transformer},
  author={Dong, Yuan and Fang, Chuan and Bo, Liefeng and Dong, Zilong and Tan, Ping},
  booktitle={CVPR},
  year={2024}
}

@inproceedings{huang2025midi,
  title={Midi: Multi-instance diffusion for single image to 3d scene generation},
  author={Huang, Zehuan and Guo, Yuan-Chen and An, Xingqiao and Yang, Yunhan and Li, Yangguang and Zou, Zi-Xin and Liang, Ding and Liu, Xihui and Cao, Yan-Pei and Sheng, Lu},
  booktitle={CVPR},
  year={2025}
}

@article{meng2025scenegen,
  title={Scenegen: Single-image 3d scene generation in one feedforward pass},
  author={Meng, Yanxu and Wu, Haoning and Zhang, Ya and Xie, Weidi},
  journal={arXiv},
  year={2025}
}

@inproceedings{li2024dreamscene,
  title={Dreamscene: 3d gaussian-based text-to-3d scene generation via formation pattern sampling},
  author={Li, Haoran and Shi, Haolin and Zhang, Wenli and Wu, Wenjun and Liao, Yong and Wang, Lin and Lee, Lik-hang and Zhou, Peng Yuan},
  booktitle={ECCV},
  year={2024}
}

@article{zhou2024layoutyour3d,
  title={Layout-your-3d: Controllable and precise 3d generation with 2d blueprint},
  author={Zhou, Junwei and Li, Xueting and Qi, Lu and Yang, Ming-Hsuan},
  journal={arXiv},
  year={2024}
}

@article{Zhou2024GALA3D,
  title={GALA3D: Towards Text-to-3D Complex Scene Generation via Layout-guided Generative Gaussian Splatting},
  author={Xiaoyu Zhou and Xingjian Ran and Yajiao Xiong and Jinlin He and Zhiwei Lin and Yongtao Wang and Deqing Sun and Ming-Hsuan Yang},
  journal={arXiv},
  year={2024}
}

@inproceedings{han2025reparo,
  title={Reparo: Compositional 3d assets generation with differentiable 3d layout alignment},
  author={Han, Haonan and Yang, Rui and Liao, Huan and Xing, Jiankai and Xu, Zunnan and Yu, Xiaoming and Zha, Junwei and Li, Xiu and Li, Wanhua},
  booktitle={ICCV},
  year={2025}
}

@inproceedings{ardelean2025gen3dsr,
  title={Gen3dsr: Generalizable 3d scene reconstruction via divide and conquer from a single view},
  author={Ardelean, Andreea and {\"O}zer, Mert and Egger, Bernhard},
  booktitle={3DV},
  year={2025}
}

@inproceedings{gu2025artiscene,
  title={ArtiScene: Language-Driven Artistic 3D Scene Generation Through Image Intermediary},
  author={Gu, Zeqi and Cui, Yin and Li, Zhaoshuo and Wei, Fangyin and Ge, Yunhao and Gu, Jinwei and Liu, Ming-Yu and Davis, Abe and Ding, Yifan},
  booktitle={CVPR},
  year={2025}
}

@article{dong2025hiscene,
  title={HiScene: creating hierarchical 3D scenes with isometric view generation},
  author={Dong, Wenqi and Yang, Bangbang and Yang, Zesong and Li, Yuan and Hu, Tao and Bao, Hujun and Ma, Yuewen and Cui, Zhaopeng},
  journal={arXiv},
  year={2025}
}

@article{yao2025cast,
  title={Cast: Component-aligned 3d scene reconstruction from an rgb image},
  author={Yao, Kaixin and Zhang, Longwen and Yan, Xinhao and Zeng, Yan and Zhang, Qixuan and Xu, Lan and Yang, Wei and Gu, Jiayuan and Yu, Jingyi},
  journal={TOG},
 
  year={2025}
}

@inproceedings{liu2024groundingdino,
  title={Grounding dino: Marrying dino with grounded pre-training for open-set object detection},
  author={Liu, Shilong and Zeng, Zhaoyang and Ren, Tianhe and Li, Feng and Zhang, Hao and Yang, Jie and Jiang, Qing and Li, Chunyuan and Yang, Jianwei and Su, Hang and others},
  booktitle={ECCV},
  year={2024}
}

@inproceedings{kirillov2023sam,
  title={Segment anything},
  author={Kirillov, Alexander and Mintun, Eric and Ravi, Nikhila and Mao, Hanzi and Rolland, Chloe and Gustafson, Laura and Xiao, Tete and Whitehead, Spencer and Berg, Alexander C and Lo, Wan-Yen and others},
  booktitle={ICCV},
  year={2023}
}

@article{ren2024groundedsam,
  title={Grounded sam: Assembling open-world models for diverse visual tasks},
  author={Ren, Tianhe and Liu, Shilong and Zeng, Ailing and Lin, Jing and Li, Kunchang and Cao, He and Chen, Jiayu and Huang, Xinyu and Chen, Yukang and Yan, Feng and others},
  journal={arXiv},
  year={2024}
}

@inproceedings{szymanowicz2025flash3d,
  title={Flash3d: Feed-forward generalisable 3d scene reconstruction from a single image},
  author={Szymanowicz, Stanislaw and Insafutdinov, Eldar and Zheng, Chuanxia and Campbell, Dylan and Henriques, Joao F and Rupprecht, Christian and Vedaldi, Andrea},
  booktitle={3DV},

  year={2025}
}

@article{wu2025amodal3r,
  title={Amodal3r: Amodal 3d reconstruction from occluded 2d images},
  author={Wu, Tianhao and Zheng, Chuanxia and Guan, Frank and Vedaldi, Andrea and Cham, Tat-Jen},
  journal={arXiv},
  year={2025}
}

@inproceedings{wang2025vggt,
  title={Vggt: Visual geometry grounded transformer},
  author={Wang, Jianyuan and Chen, Minghao and Karaev, Nikita and Vedaldi, Andrea and Rupprecht, Christian and Novotny, David},
  booktitle={CVPR},
  year={2025}
}

@article{keetha2025mapanything,
  title={MapAnything: Universal feed-forward metric 3D reconstruction},
  author={Keetha, Nikhil and M{\"u}ller, Norman and Sch{\"o}nberger, Johannes and Porzi, Lorenzo and Zhang, Yuchen and Fischer, Tobias and Knapitsch, Arno and Zauss, Duncan and Weber, Ethan and Antunes, Nelson and others},
  journal={arXiv},
  year={2025}
}

@inproceedings{zheng2020structured3d,
  title={Structured3d: A large photo-realistic dataset for structured 3d modeling},
  author={Zheng, Jia and Zhang, Junfei and Li, Jing and Tang, Rui and Gao, Shenghua and Zhou, Zihan},
  booktitle={ECCV},
  year={2020}
}

@inproceedings{fu20213dfront,
  title={3d-front: 3d furnished rooms with layouts and semantics},
  author={Fu, Huan and Cai, Bowen and Gao, Lin and Zhang, Ling-Xiao and Wang, Jiaming and Li, Cao and Zeng, Qixun and Sun, Chengyue and Jia, Rongfei and Zhao, Binqiang and others},
  booktitle={ICCV},
  year={2021}
}

@inproceedings{suvorov2022lama,
  title={Resolution-robust large mask inpainting with fourier convolutions},
  author={Suvorov, Roman and Logacheva, Elizaveta and Mashikhin, Anton and Remizova, Anastasia and Ashukha, Arsenii and Silvestrov, Aleksei and Kong, Naejin and Goka, Harshith and Park, Kiwoong and Lempitsky, Victor},
  booktitle={WACV},
  year={2022}
}

@article{feng2025dit360,
  title={DiT360: High-Fidelity Panoramic Image Generation via Hybrid Training},
  author={Feng, Haoran and Zhang, Dizhe and Li, Xiangtai and Du, Bo and Qi, Lu},
  journal={arXiv},
  year={2025}
}

@article{wang2024depthanywhere,
  title={Depth anywhere: Enhancing 360 monocular depth estimation via perspective distillation and unlabeled data augmentation},
  author={Wang, Ning-Hsu Albert and Liu, Yu-Lun},
  journal={NIPS},
  year={2024}
}

@article{hu2025flashsculptor,
  title={Flash Sculptor: Modular 3D Worlds from Objects},
  author={Hu, Yujia and Liu, Songhua and Yang, Xingyi and Wang, Xinchao},
  journal={arXiv},
  year={2025}
}

@ARTICLE{besl1992icp,
  title={Method for registration of 3-D shapes},
  author={Besl, Paul J and McKay, Neil D},
  journal={TPAMI}, 
  year={1992}
}

@inproceedings{yang2024holodeck,
  title={Holodeck: Language guided generation of 3d embodied ai environments},
  author={Yang, Yue and Sun, Fan-Yun and Weihs, Luca and VanderBilt, Eli and Herrasti, Alvaro and Han, Winson and Wu, Jiajun and Haber, Nick and Krishna, Ranjay and Liu, Lingjie and others},
  booktitle={CVPR},
  year={2024}
}

@article{feng2023diffusion360,
  title={Diffusion360: Seamless 360 degree panoramic image generation based on diffusion models},
  author={Feng, Mengyang and Liu, Jinlin and Cui, Miaomiao and Xie, Xuansong},
  journal={arXiv},
  year={2023}
}

@article{small_gicp,
author = {Kenji Koide},
title = {{small\_gicp: Efficient and parallel algorithms for point cloud registration}},
journal = {Journal of Open Source Software},
year = {2024}
}

@article{ho2020diffusion,
  title={Denoising diffusion probabilistic models},
  author={Ho, Jonathan and Jain, Ajay and Abbeel, Pieter},
  journal={NIPS},
  year={2020}
}

@article{oquab2023dinov2,
  title={Dinov2: Learning robust visual features without supervision},
  author={Oquab, Maxime and Darcet, Timoth{\'e}e and Moutakanni, Th{\'e}o and Vo, Huy and Szafraniec, Marc and Khalidov, Vasil and Fernandez, Pierre and Haziza, Daniel and Massa, Francisco and El-Nouby, Alaaeldin and others},
  journal={arXiv},
  year={2023}
}

@inproceedings{zhang2024ram,
  title={Recognize anything: A strong image tagging model},
  author={Zhang, Youcai and Huang, Xinyu and Ma, Jinyu and Li, Zhaoyang and Luo, Zhaochuan and Xie, Yanchun and Qin, Yuzhuo and Luo, Tong and Li, Yaqian and Liu, Shilong and others},
  booktitle={CVPR},
  year={2024}
}
